\documentclass{article}

 \usepackage[final,nonatbib]{nips_2016}

\usepackage[utf8]{inputenc} 
\usepackage[T1]{fontenc}    
\usepackage{hyperref}       
\usepackage{url}            
\usepackage{booktabs}       
\usepackage{amsfonts}       
\usepackage{nicefrac}       
\usepackage{microtype}      
\usepackage{graphicx}
\usepackage{amsmath}
\usepackage{yhmath}

\title{Leveraging Recurrent Neural Networks for Multimodal Recognition of Social Norm Violation in Dialog}

\author{
  Tiancheng Zhao$^{1}$, Ran Zhao$^{1}$, Zhao Meng$^{2}$, Justine Cassell$^{1}$\\
  $^{1}$Language Technologies Institute, Carnegie Mellon University, Pittsburgh, USA\\
  $^{2}$Department of Computer Science, Peking University, Beijing, China\\
\texttt{\{tianchez,rzhao1,Justine\}}@cs.cmu.edu \\
  \texttt{zhaomeng.pku@outlook.com} \\
}

\begin{document}

\maketitle

\begin{abstract}
Social norms are shared rules that govern and facilitate social interaction. Violating such social norms via teasing and insults may serve to upend power imbalances or, on the contrary reinforce solidarity and rapport in conversation, rapport which is highly situated and context-dependent. In this work, we investigate the task of automatically identifying the phenomena of social norm violation in discourse. Towards this goal, we leverage the power of recurrent neural networks and multimodal information present in the interaction, and propose a predictive model to recognize social norm violation. Using long-term temporal and contextual information, our model achieves an F1 score of 0.705. Implications of our work regarding developing a social-aware agent are discussed.    
\end{abstract}
\section{Introduction and Related Work}
Social norms are informal understandings that govern human behavior. They serve as the basis for our beliefs and expectations about others, and are instantiated in human-human conversation through verbal and nonverbal behaviors \cite{boella2003obligations,traum1994discourse}. There is considerable body of work on modeling socially normative behavior in intelligent agent-based systems \cite{si2006thespian,paiva2004caring}, aiming to facilitate lifelike conversations with human users. Violating such social norms and impoliteness in the conversation, on the other hand, have also been demonstrated to positively affect certain aspects of the social interaction. For instance, \cite{52} suggests impoliteness may challenge rapport in strangers but it is also an indicator of built relationship among friends. The literature on social psychology  \cite{spencer2005politeness} shows that the task of managing interpersonal bond like rapport requires management of face which, in turn, relies on behavioral expectation, which are allied with social norms early in a relationship, and become more interpersonally determined as the relationship proceeds. \cite{25} advanced the arguments by proposing that with the increasing knowledge of one another, more general norms may be purposely violated in order to accommodate each other's behavior expectation. Moreover, they proposed that such kind of social norm violation in fact reinforce the sense of in-group connectedness. Finally in \cite{ranIva2016}, the authors discovered the effect of temporally co-occurring smile and social norm violation that signal high interpersonal rapport. Thus, we believe that recognizing the phenomena of social norm violation in dialog can contribute important insights into understanding the interpersonal dynamics that unfold between the interlocutors.

Interesting prior work on quantifying social norm violation has taken a heavily data-driven focus \cite{danescu2013no,wang2016modeling}. For instance, \cite{danescu2013no} trained a series of bigram language models to quantify the violation of social norms in users' posts on an online community by leveraging cross-entropy value, or the deviation of word sequences predicted by the language model and their usage by the user. However, their models were trained on written-language instead of natural face-face dialog corpus. Another kind of social norm violation was examined by \cite{riloff2013sarcasm}, who developed a classifier to identify specific types of sarcasm in tweets. They utilized a bootstrapping algorithm to automatically extract lists of positive sentiment phrases and negative situation phrases from given sarcastic tweets, which were in turn leveraged to recognize sarcasm in an SVM classifier. However, no contextual information was considered in this work. \cite{ran2016sigdial} understood the nature of social norm violation in dialog by correlating it with associated observable verbal, vocal and visual cues. By leveraging their findings and statistical machine learning techniques, they built a computational model for automatic recognition. While they preserved short-term temporal contextual information in the model, this study avoided dealing with sparsity of the social norm violation phenomena by under-sampling the negative-class instances to make a balanced dataset.

Motivated by theoretical rationale and prior empirical findings concerning the relationship between violation social norm and interpersonal dynamics, in the current work, we take a step towards addressing the above limitations and our contributions are two-fold: (1)We quantitatively evaluate the contribution of long-term temporal contextual information on detecting violation of social norm. (2)We incorporate this understanding to our computational model for automatic recognizing social norm violation by leveraging the power of recurrent neural network on modeling the long-term temporal dependencies. 
\section{Data and Annotation}
Reciprocal peer tutoring data was collected from 12 American English-speaking dyads (6 friends and 6 strangers; 6 boys and 6 girls), with a mean age of 13 years, who interacted for 5 hourly sessions over as many weeks (a total of 60 sessions, and 5400 minutes of data), tutoring one another in algebra. Each session began with a period of getting to know one another, after which the first tutoring period started, followed by another small social interlude, a second tutoring period with role reversal between the tutor and tutee, and then the final social time. 

We assessed our automatic recognition of social norm violation against this corpus annotated for those strategies. Inter-rater reliability (IRR) for the social norm violation that computed via Krippendorff's alpha was 0.75. IRR for visual behavior was 0.89 for eye gaze, 0.75 for smile count (how many smiles occur), 0.64 for smile duration and 0.99 for head nod. Table~\ref{tab:statistics} shows statistics of our corpus. Below we discuss the definition of social norm violation. 
\begin{table}[h!]
\centering
\caption{Statistics of the corpus}
\label{tab:statistics}
\begin{tabular}{l|l|l|l} \hline
      & Train & CV & Test \\ \hline
\#clauses & 39,254  & 5,491 & 7,008          \\
\#sessions &  48  &   6 &  6
 \\\hline
\end{tabular}
\end{table}
\\
\textbf{Ground Truth:} Social norm violations are behaviors or actions that go against general socially acceptable and stereotypical behaviors. In a first pass, we coded whether a clause was a social norm violation. In a second pass, if a social norm violation, we differentiated: (1) breaking the conversational rules of the experiment (e.g. off-task talk during tutoring session, insulting the experimenter or the experiment, etc); (2) face threatening acts (e.g. criticizing, teasing, or insulting, etc); (3) referring to one's own or the other person's social norm violations or general social norm violations (e.g. referring to the need to get back to focusing on work, or to the other person being verbally annoying etc). Social norms are culturally-specific, and so we judged a social norm violation by the impact it had on the listener (e.g. shock, specific reference to the behavior as a violation, etc.).
\section{Model and Experiment}
In this section, our objective was to build a computational model for detecting social norm violation. Towards this end, we first took each clause, the smallest units that can express a complete proposition, as the prediction unit. Next, inspired from the thorough analysis in \cite{ran2016sigdial}, we extracted verbal and visual features of the speaker that were highly correlated to social norm violation clauses, with rare threshold being set to 20. Verbal features included LIWC features\cite{pennebaker2015development} that helped in categorization of words used during usage of social norm violation, bigrams, part of speech bigrams and word-part of speech pairs from the speaker's clauses. Visual features included head node, smile and eye gaze information of the speaker. In total there were 3782 features per clause.
\subsection{Models}
We treated a dialog $D$ as a sequence of clauses $c_0, ... c_T$, where $T$ was the number of clauses in the $D$. Each clause $c_i$ was a tuple $([w^i_0, ...w^i_m], e_i)$, where $[w^i_0, ...w^i_m]$ was the $m$ words in the clause $c_i$, and $e_i$ was the corresponding meta information such as the relationship of the dyad and nonverbal behavior during the generation of the clause. The handcrafted feature of size 3782 was denoted as $f_i$, and could be viewed as a mapping function $F: c_i \rightarrow f_i$.  Meanwhile, each clause was associated with a binary label $y_i \in \{0, 1\}$ that indicates the ground truth of whether $c_i$ is a violation of social norm. Eventually, the goal was to model $p(y_t|c_{0:t})$, the conditional distribution over whether the latest clause was a violation of social norm, given the entire history of the dialog.
\subsubsection{Logistic Regression Model}
We first trained a L2 regularized logistic regression model using the proposed verbal and visual features $f_i$ as inputs (leftmost in Figure 1). This model serves as our baseline.

\subsubsection{Local/Global-Context RNN Model}
\begin{figure}[h!]
	\centering
	\includegraphics[width=1\textwidth]{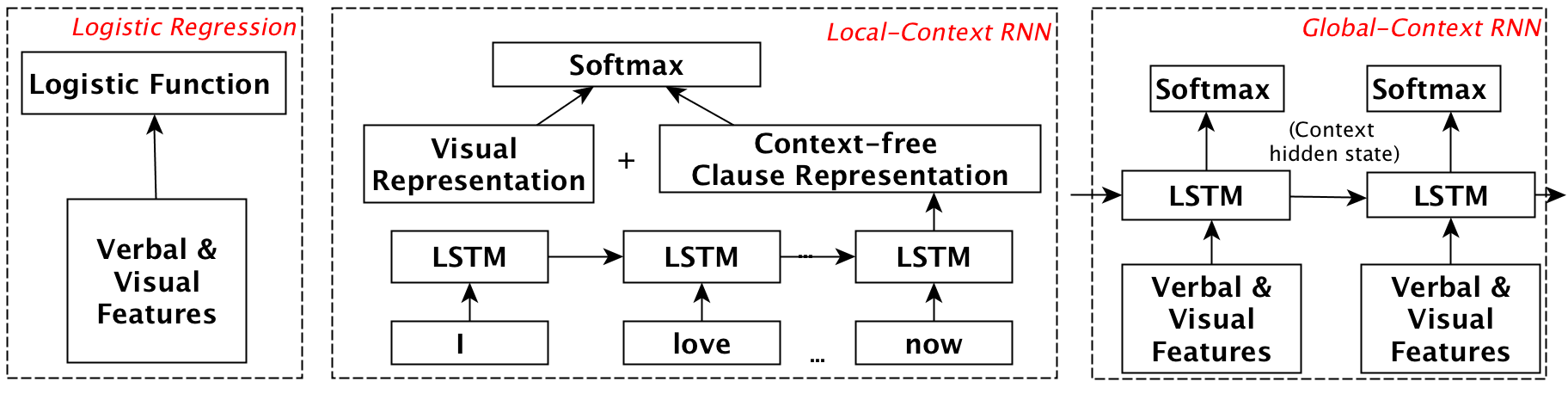}
	\caption{Three proposed computational models.}
	\label{fig:model2}
\end{figure}
Past empirical results suggest two possible hypotheses of improving the model performance: 1. improvement in clause level representation 2. inclusion of contextual information for prediction. Therefore, we designed Local/Global-Context models to test these hypotheses. 

The Local-Context recurrent neural network (RNN) models the context inside a clause at the word-level by encoding word embeddings of size 300 in a clause $c_i$ sequentially using a Long-short Term Memory (LSTM) cell of size 300. The mechanism of LSTM is defined as: 
\begin{align*}
  \left[
    \begin{matrix}
      i_t  \\
      f_t  \\
      o_t  \\
      j_t  \\
    \end{matrix}
\right] &= 
  \left[
    \begin{matrix}
      \sigma  \\
      \sigma  \\
      \sigma  \\
      tanh  \\
    \end{matrix}
\right] W [h_{t-1}, x_t]  \\
    c_t &= f_t \odot c_{t-1} + i_t \odot j_t\\
    h_t &= o_t \odot tanh(c_t)
\end{align*}
We treated last hidden LSTM output $h^i_m$ as the clause embedding and concatenated that with the corresponding meta information vector $e_i$. The combined vector was linearly transformed and then fed into a softmax function. 

Next our Global-Context RNN investigated the influence of clause-level context in detecting social norm violation, by using the LSTM cells to model the long-term temporal dependencies. For a fair comparison, we used the same hand-crafted feature $f_i$ used in the logistic regression model as the representation of clause $c_i$. As shown in Figure~\ref{fig:model2}, we first obtained a linear embedding of size 150 $emb_i=W_{e}f_i+b_i$ of $f_i$. Then $emb_i$ was used as the inputs to LSTM of size 600. The hidden output $h_i$ at each time step was fed into a multilayer perceptron (MLP) with 1 hidden layer of size 100. We applied 50\% dropout regularization~\cite{zaremba2014recurrent} at the input/output of LSTM and MLP hidden layer for better generalization. Finally the model was optimized w.r.t to the cross entropy loss. A further challenge was the length of dialog. The average number of clauses in training dialog was 817.8, which made it computationally intractable to backpropagate through the entire sequence. Therefore, truncated backpropagation through time (TBPTT)~\cite{sutskever2013training} was used by unrolling the network for 20 steps. The final state of LSTM of each batch was fetched into the next batch as the initial state.
\subsection{Experiment Result}
We observed that Global-Context RNN with 2 LSTM layers outperformed other models as showed in Table 2. First, by comparing logistic regression model with our best model, the result indicates the strong predictive power of long-term temporal contextual information on the task of detecting social norm violation in dialog. On the other hand, Local-Context RNN model did not achieve significant improvement on overall performance regarding to logistic regression, which means that our learned clause representation through training process has less competence compared to hand-crafted features inspired from linguistic knowledge. One potential reason for such a result could be insufficient amount of training set in order to learn a generic clause representation.
\begin{table}[h!]
\centering
\caption{Performance comparsion for the 3 evaluated models}
\label{tbl:results}
\begin{tabular}{l|l|l|l} \hline
      & Precision & Recall & F-measure \\ \hline
Logistic Regression & 0.573  &0.583 & 0.578          \\
Local-Context RNN   &  0.478  &   0.747 &  0.583\\
Global-Context RNN (1-layer)  & 0.689 &  0.696 & 0.693\\
Global-Context RNN (2-layer)  & \textbf{0.690}  & \textbf{0.720} & \textbf{0.705} \\\hline
\end{tabular}
\end{table}
\section{Conclusion and Future Work}
In this work, we began by indicating our interest in quantitatively learning the contribution of long-term temporal contextual information on detecting social norm violation in discourse. We then leveraged the power of recurrent neural network on modeling long-term temporal dependency. Inspired by hand-crafted multimodal features derived from qualitative and quantitative analysis in former empirical studies, we developed a Global-Context RNN model to detect social norm violation in human dialog. This model will play a prime role in building socially-aware agents that have capabilities of understanding interpersonal dynamics that unfold in the interaction, which is in turn, essential to better adapt to the interpersonal relationship felt by their users. Thus, to serve this goal, our future work will build a generative model of social norm violation, which will make an agent act towards more realistic human behavior understanding, reasoning and generation. We begin to model those aspects of human-human interaction that are not only helpful to human-agent collaboration, but also sustain aspects of what we cherish most in being human.
\bibliography{nips_2016}
\bibliographystyle{plain}  
\end{document}